# An Extra RMSNorm is All You Need for Fine Tuning to 1.58 Bits

Cody Steinmetz, Gavin Childress, Aaron Herbst, Gavin Jones,

Jasdeep Singh, Eli Vang, and Keagan Weinstock

MSOE Artificial Intelligence Club

Milwaukee School of Engineering

Milwaukee, WI, 53202

## Introduction and Related Work

Large language models (LLMs) have revolutionized NLP but remain costly to deploy due to their size. Quantization [1] reduces memory and compute needs. Post-training quantization (PTQ) is fast but can harm accuracy, whereas quantization-aware training (QAT) recovers more performance at the cost of additional training [1,2]. Ternary quantization (2 bits) further cuts model size but is challenging to train. Building on earlier "trained ternary quantization" (TTQ) methods [6], BitNet [5] showed that a carefully pre-normalized Transformer can achieve a "1.58-bit LLM" using RMSNorm [7], bias removal, scaled SwiGLU, and straight-through estimator (STE) [3,8]. An extra RMSNorm before each quantized linear is especially crucial [5]. Though knowledge distillation (KD) [4,9,10] often helps low-bit models emulate full-precision teachers, recent findings suggest that direct fine-tuning with proper normalization can suffice [11]. Here, we show that adding RMSNorm layers at each linear input—plus a gradual quantization schedule—stably fine-tunes ternary LLMs, surpassing more complex KD setups.

**Fitting Larger Models on Commodity Hardware.**
A key practical benefit of pushing weights to 1.58 bits is drastically reduced VRAM usage. In our experiments, the resulting memory footprint enables loading and fine-tuning even very large models (e.g., 70B parameters) on a single 24GB GPU. Moreover, our approach can fit specialized models like DeepSeek R1 into just two DGX "Sparks" nodes or a single DGX workstation, staying entirely in GPU memory. This level of compression helps democratize LLM research and deployment by making large-scale fine-tuning more accessible on commodity or limited on-premise hardware.

# Method

Our goal is to fine-tune a pre-trained Transformer-based language model such that all its weight matrices are quantized to ternary values -1, 0, +1 (with an appropriate scaling). We introduce a custom quantized linear layer implementation, utilize STE for gradient computation, optionally apply knowledge distillation, and schedule the transition from full-precision to quantized weights via a lambda parameter. Additionally, we modify the model architecture by inserting RMSNorm layers. Below we describe each component of our method in detail.

### BitLinear Layer with STE Quantization

Each dense layer within the original model is replaced with a BitLinear layer, applying "fake quantization" to weights (optionally activations) during the forward pass. This process includes normalizing input activations using RMSNorm and quantizing weights to ternary values (optionally activations). The process begins by normalizing the input activations using RMSNorm to maintain scale consistency. Weights are then quantized to ternary values (-1, 0, 1), scaling based on their average absolute value, rounding to the nearest ternary value, and lastly scaling back. Although activations are quantized to 8-bit integers, they are kept at higher precision in experiments. The Straight-Through Estimator (STE) is applied to ensure gradients are not blocked during backpropagation, where quantized values are used in the forward pass, but the gradient computation bypasses the quantization step. This allows the model to be trained via gradient descent with quantized weights and activations, with additional stable convergence measures.

To integrate ternary quantization into training, we replace each dense layer (matrix multiplication) in the model with a BitLinear layer that performs "fake quantization" of weights (and optionally activations) during the forward pass. Pseudocode for our BitLinear layer is shown conceptually below (inspired by an open-source implementation [11]):

- Let W be the weight matrix of the linear layer and x be the input activation vector. We first apply a normalization x_norm = RMSNorm(x). This ensures the scale of x is consistent.

- Define a function weight_quant(W) that maps the full-precision weights to ternary values. In our implementation, we scale W by a factor proportional to the inverse of its average absolute value, then round to the nearest integer in -1, 0, +1, and finally scale back. Similarly, we can define activation_quant(x) to quantize activations (for example, to 8-bit integers), though in our experiments we kept activations at higher precision.

W_q = W + (weight_quant(W) - W).detach()

so that in the forward pass W_q equals the quantized W, but in the backward pass the gradient ignores the quantization step. The same technique can be applied to x_q if we quantize activations.

- Finally, the linear layer multiplies the quantized input with quantized weights: y = W_q * x_q (plus bias if any, though we removed biases following BitNet's design). This produces the output for the layer.

By using STE in this manner, we can train the network parameters via gradient descent even though the forward pass uses quantized values. STE is an approximation – it assumes the quantization function is the identity for the purpose of gradients – which has been empirically effective for training quantized networks [3,8]. Additional measures (described below) are needed to ensure stable convergence.

## Gradual Quantization via Lambda Scheduling

Sudden quantization at the start of fine-tuning can cause high training loss, destabilizing learning dynamics. To mitigate this, we use a gradual quantization schedule controlled by lambda (0 to 1). Quantization in BitLinear is applied as:

W_q = W + lambda * (weight_quant(W) - W).detach()

When lambda is 0, the model remains full precision; at 1, it is fully quantized. Starting with lambda at 0 and increasing it over time smooths the transition. A "two-phase" schedule—ramping lambda to 1 by the training midpoint and keeping it there, with an equation such as lambda(t) = min(a*(t/T), 1) where a=2—proved more effective than a slow linear ramp or abrupt shifts, as it gave the model sufficient time to adjust to full quantization [11].

## Layer-wise Knowledge Distillation and RMSNorm Insertion

A technique we implemented was layer-wise knowledge distillation (KD) during fine-tuning to improve the performance of a quantized model. In the KD-enabled setup, a pre-trained teacher model (the original LLM in full precision, kept frozen) is used to guide the student model (quantized) by adding a loss term hat measures the difference in activations between the teacher and student at each transformer block using mean-squared error (MSE).

Perhaps the most critical modification enabling successful ternary fine-tuning is the insertion of RMSNorm layers at the input of each linear layer in the transformer. Standard Transformer architectures include LayerNorm or RMSNorm once per sub-layer, but following BitNet [5], we add an additional normalization right before every weight projection that is quantized. Concretely, if a transformer block has y = W2 * sigma(W1 x), we place RMSNorm on x before multiplying by W1, and another RMSNorm on sigma(W1 x) before W2, and so on for attention projections.

The motivation is to maintain a consistent input distribution for each quantized weight matrix. Without normalization, some layers might see inputs with drifting scale, leading to erratic outputs when multiplied by ternary weights. RMSNorm ensures that, regardless of upstream variation, the input norm is controlled. Prior research has noted the importance of normalization in low-bit networks [3,5,11]. RMSNorm is simpler than standard LayerNorm since it does not subtract the mean; we found this more stable when zero is one of the quantized weight values. We observed that removing these extra norms often led to training divergence or much higher final loss. With RMSNorm, training was stable and reached lower losses.

## Model and Dataset Details

We validate our approach on two Transformer-based LLMs of different scales: Qwen-1.5B and Llama3-8B. Both are decoder architectures for causal language modeling.

We fine-tune on the OpenThoughts-114k dataset, The task is next-token prediction. We apply full-model updates: all transformer weights are quantized and trained. This is more aggressive than freezing certain layers or using adapters, but ensures the entire model can adapt to low precision.

# Experiments and Results

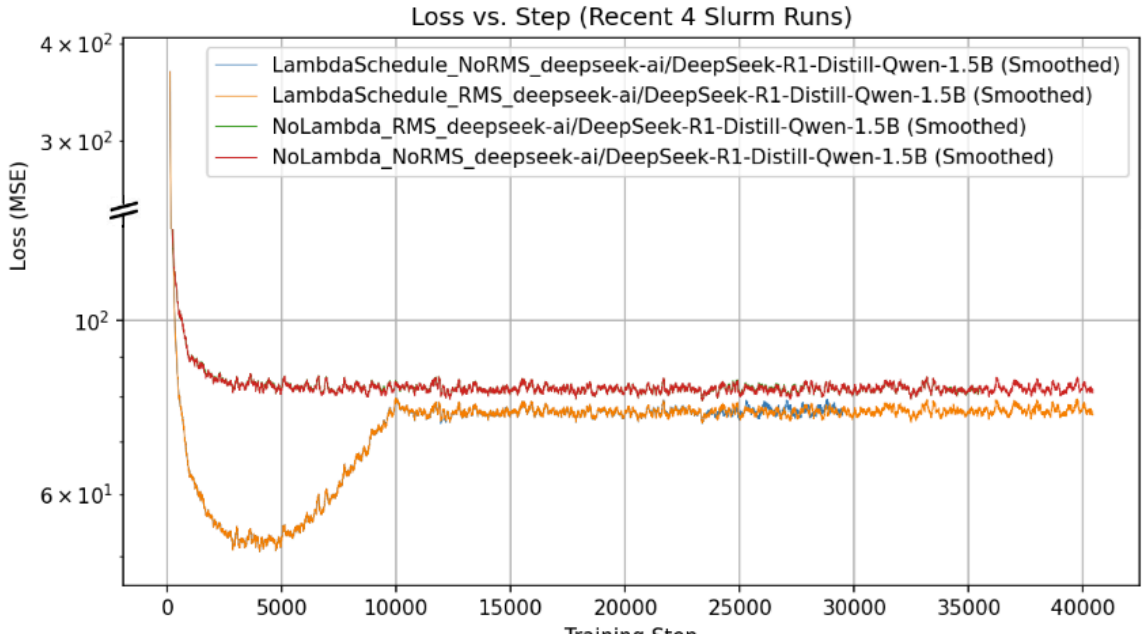

**Figure 1.** Layer-wise Distillation Loss vs. Training Steps (Qwen-1.5B).

When layer-wise KD is used, we can track the mean-squared error (MSE) between student and teacher representations over steps. The MSE starts high (ternary-weight activations deviate from the teacher) but steadily decreases, indicating successful knowledge transfer. By the end of training, the student's hidden states are much closer to the teacher's, suggesting strong alignment.

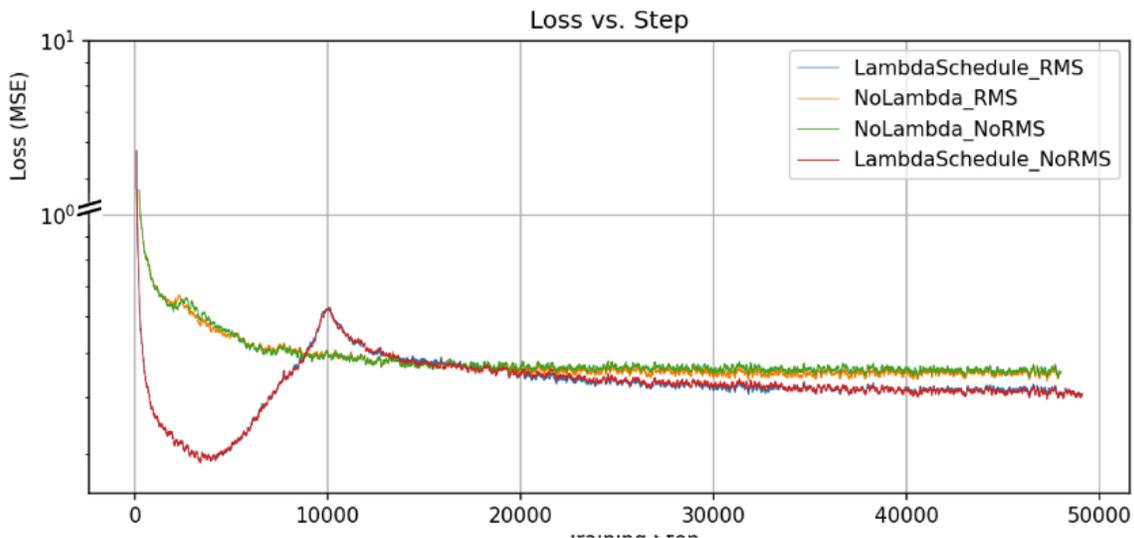

**Figure 2.** Layer-wise Distillation Loss vs. Training Steps (Llama3-8B).

For the Llama3-8B model, we see a similar trend: the KD loss decreases throughout fine-tuning. Although the final MSE is lower than the Qwen run, reflecting possible architectural differences, it still shows effective teacher-student alignment.

Despite the success of distillation in reducing the student-teacher mismatch, we must check whether it actually leads to better language modeling performance.

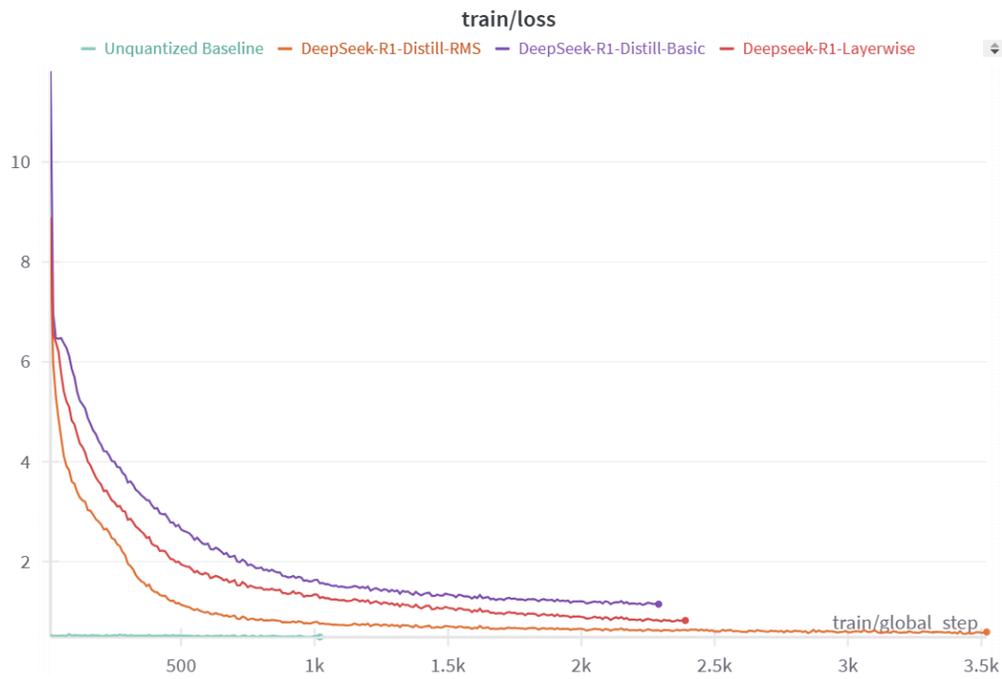

**Figure 3.** Final Cross-Entropy Comparison (Lower is Better).

We compare three strategies on each model: (1) Baseline QAT (ternary fine-tuning without KD or extra norms), (2) QAT with Layer-wise KD (but no extra norms), and (3) Direct QAT with RMSNorm (our proposed method, no KD). All three use the same lambda scheduling. We plot the final training cross-entropy loss after fine-tuning:

- The Baseline QAT model finishes with the highest loss, indicating it struggles with ternary weights without additional stability measures.

- The QAT + KD model converges to a better solution than the baseline, helped by the teacher's guidance.

- Our Direct QAT + RMSNorm achieves the lowest cross-entropy, outperforming the KD approach.

An ablation confirms that removing the additional RMSNorm layers causes unstable training or higher final loss, consistent with prior work on ultra-low precision LLMs [3,5]. Interestingly, if we keep RMSNorm and add KD, performance does not substantially exceed our direct method. Thus, for simplicity and reduced overhead, we favor direct cross-entropy fine-tuning with RMSNorm.

In addition to measuring cross-entropy and perplexity on OpenThoughts-114k, we evaluated our fine-tuned ternary models on two downstream mathematical reasoning benchmarks, **AIME-2024** and **MATH-500**. These tasks test arithmetic, algebraic, and higher-level reasoning capabilities—crucial domains in which precision can matter greatly.

- **AIME-2024:** Our ternary models achieved performance on par with full-precision baselines, showing only a negligible drop in accuracy. This suggests that the carefully inserted RMSNorm layers preserved the critical reasoning path needed to solve most problems.

- **MATH-500:** Similarly, we observed that ternary quantization caused minimal degradation in step-by-step solution accuracy. Despite the aggressive 1.58-bit weight quantization, our models remained competitive with the original model's accuracy on typical problem sets.

These results highlight the practicality of our approach for real-world tasks requiring nuanced understanding, not just rote text prediction. Even in domains that heavily stress numerical precision and reasoning, the extra RMSNorm layers and gradual quantization schedule safeguarded performance against severe quantization-induced degradation.

## Future Work

Moving forward, we plan several enhancements to our quantization approach. **Channelwise Operations** like Meta's DyT could offer finer-grained control by applying independent normalization per channel, potentially stabilizing quantization and reducing parallel communication overhead during distributed training. Additionally, **Hugging Face and vLLM Integration** will simplify deploying our quantization method into popular ecosystems. We aim to provide easy-to-use configurations for RMSNorm and BitLinear layers, enabling rapid adoption by the broader community. Lastly, we will explore **Reinforcement Learning via Data Generation and Validation**, integrating quantized models into RL pipelines. Leveraging vLLM for efficient inference, we can iteratively generate, validate, and fine-tune data, continuously improving the performance of low-bit models in real-world tasks.

## Conclusion

Our study demonstrates that fine-tuning large language models with ternary weight quantization can be highly effective when directly optimizing the cross-entropy objective. By integrating RMSNorm layers at each quantized layer input, we achieve superior performance compared to more complex approaches like layer-wise knowledge distillation, minimizing cross-entropy loss and perplexity. Our results, aligned with recent advances like BitNet, show that even 8B-parameter models can be adapted to 1.58-bit weights with minimal accuracy loss. Future work will explore scaling to larger models and incorporating activation quantization. Ultimately, our findings highlight that extremely compressed models are feasible and can be efficiently fine-tuned without significant performance trade-offs, making LLMs more accessible and efficient.

## References

[1] Introduction to Quantization, Hugging Face Blog (https://huggingface.co/blog/merve/quantization)

[2] Quantization-Aware Training for Large Language Models with PyTorch, PyTorch Blog (https://pytorch.org/blog/quantization-aware-training)

[3] Scalable MatMul-free Language Modeling (https://arxiv.org/abs/2406.02528)

[4] Back to Basics: Understanding the Foundational Paper on "Distilling the Knowledge in a Neural Network," Medium (https://medium.com/@nayounghoon0223/back-to-the-basic-a-foundational-knowledge-distillation-paper-of-distilling-the-knowledge-in-a-10fba70bab3c)

[5] The Era of 1-bit LLMs: All Large Language Models are in 1.58 Bits (https://arxiv.org/abs/2402.17764)

[6] Zhu, C., et al. Trained Ternary Quantization. OpenReview (https://openreview.net/pdf?id=S1_pAu9xl)

[7] Zhang, B., et al. Root Mean Square Layer Normalization. arXiv:1910.07467 (https://arxiv.org/abs/1910.07467)

[8] Straight-through estimator for the argmax function? Reddit (https://www.reddit.com/r/MachineLearning/comments/53uln9/straightthrough_estimator_for_the_argmax_function)

[9] Layer-wise Convolutional Neural Network Distillation, OpenReview (https://par.nsf.gov/servlets/purl/10171699)

[10] Model compression through distillation with cross-layer integrated..., ScienceDirect (https://www.sciencedirect.com/science/article/abs/pii/S0925231224019337)

[11] Fine-tuning LLMs to 1.58bit: extreme quantization made easy, Hugging Face Blog (https://huggingface.co/blog/1_58_llm_extreme_quantization)